\documentclass[10pt,twocolumn,letterpaper]{article}

\usepackage{cvpr}              

\definecolor{cvprblue}{rgb}{0.21,0.49,0.74}
\usepackage[pagebackref,breaklinks,colorlinks,allcolors=cvprblue]{hyperref}

\usepackage{multirow}

\usepackage{xcolor,colortbl}

\definecolor{Gray}{gray}{0.9}
\newcolumntype{a}{>{\columncolor{Gray}}c}

\definecolor{blush}{rgb}{0.87, 0.36, 0.51}
\newcommand\best[1]{\textbf{\color{blush}#1}}
\newcommand\bestclean[1]{\textbf{#1}}

\usepackage{bbding}
\usepackage{algorithm,algorithmic}

\def\paperID{5134} 
\def\confName{CVPR}
\def\confYear{2025}

\title{R-TPT: Improving Adversarial Robustness of Vision-Language Models through Test-Time Prompt Tuning}

\author{Lijun Sheng$^{1,2}$, Jian Liang$^{2,3}$\thanks{To whom correspondence should be addressed.}\;, Zilei Wang$^{1}$, Ran He$^{2,3}$ \\
$^{1}$ University of Science and Technology of China \\
$^{2}$ NLPR \& MAIS, Institute of Automation, Chinese Academy of Sciences \\
$^{3}$ University of Chinese Academy of Sciences \\
{\tt\small slj0728@mail.ustc.edu.cn, liangjian92@gmail.com}}

\begin{document}
\maketitle

\begin{abstract}
Vision-language models (VLMs), such as CLIP, have gained significant popularity as foundation models, with numerous fine-tuning methods developed to enhance performance on downstream tasks.
However, due to their inherent vulnerability and the common practice of selecting from a limited set of open-source models, VLMs suffer from a higher risk of adversarial attacks than traditional vision models.
Existing defense techniques typically rely on adversarial fine-tuning during training, which requires labeled data and lacks of flexibility for downstream tasks.
To address these limitations, we propose robust test-time prompt tuning (R-TPT), which mitigates the impact of adversarial attacks during the inference stage.
We first reformulate the classic marginal entropy objective by eliminating the term that introduces conflicts under adversarial conditions, retaining only the pointwise entropy minimization.
Furthermore, we introduce a plug-and-play reliability-based weighted ensembling strategy, which aggregates useful information from reliable augmented views to strengthen the defense.
R-TPT enhances defense against adversarial attacks without requiring labeled training data while offering high flexibility for inference tasks.
Extensive experiments on widely used benchmarks with various attacks demonstrate the effectiveness of R-TPT.
The code is available in \url{https://github.com/TomSheng21/R-TPT}.
\end{abstract}

\section{Introduction}
Vision-language models (VLMs) \cite{radford2021learning, chen2023vlp, zhang2024vision} are multimodal models pretrained on large-scale paired image-text data.
Their powerful zero-shot inference capability and broad applicability across a range of downstream tasks have made them a foundational tool in the research community.
CLIP \cite{radford2021learning}, a milestone work, aligns features from the text and visual modalities using two specialized feature extractors trained with a contrastive loss function.
Due to its concise architecture and impressive performance, CLIP has become the most widely used VLM across diverse research topics.
For classification tasks, CLIP extracts features from both images and category descriptions, then chooses the category whose features exhibit the highest similarity to the image’s feature representation.
Beyond classification \cite{zhou2022learning, zhou2022conditional, wang2023improving, wang2024hard}, CLIP has been successfully applied to various vision tasks, such as semantic segmentation \cite{zhou2022extract, shin2022reco}, object detection \cite{zhao2022exploiting, zhang2023simple}, and image clustering \cite{cai2023semantic, li2023image}.

\begin{figure}[t]
    \centering
    \includegraphics[width=1.0\linewidth]{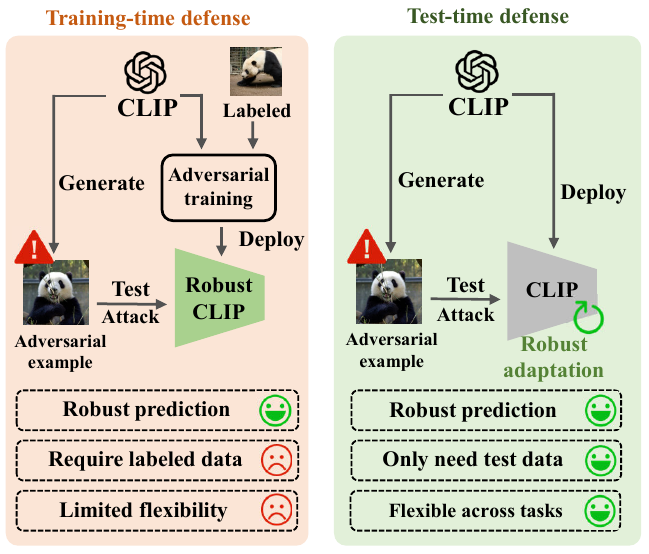}
    \caption{Comparison between training-time and test-time defense for CLIP. Our test-time defense paradigm provides robust prediction as the conventional training-time methods and requires no annotated dataset or adversarial training.}
    \label{fig:task}
\end{figure}

The impressive performance on downstream tasks and the broad range of applications of CLIP not only highlight its powerful capabilities but also expose it to potential vulnerabilities, particularly under adversarial attacks \cite{mao2022understanding, kong2024patch}.
While adversarial attacks and defenses \cite{madry2017towards, goodfellow2014explaining, zhang2019theoretically} have been widely explored for conventional visual models, the situation for CLIP is more complex.
The pre-training of CLIP requires the collection of vast amounts of image-text pairs and substantial computational resources that most deployers cannot afford.
As a result, many deployers choose to adopt open-source versions of CLIP from a small range of candidates.
This fact introduces a relatively high risk of adversarial attacks targeting CLIP-based applications.

Recent works \cite{mao2022understanding, li2024one} explore adversarial prompt tuning using annotated data to enhance the robustness of CLIP.
However, their reliance on labeled data and limited flexibility across tasks pose challenges for real-world deployment.
To address this, we choose to defend adversarial attacks in the inference stage, which is applicable across various scenarios and requires no labeled dataset or prior knowledge of the downstream task, as shown in Figure~\ref{fig:task}.
Existing test-time adaptation methods \cite{shu2022test, zanella2024test} primarily focus on improving accuracy for clean test samples, while overlooking the potential risks posed by adversarial attacks.
Moreover, deploying defense at test time necessitates short inference time and avoiding using additional resources, such as large language models or diffusion models, to ensure versatility.

To address the above challenges and achieve successful defense against potential attacks, we propose robust test-time prompt tuning (R-TPT).
First, we revisit and refine the widely used optimization objective for instance adaptation.
Many previous works \cite{shu2022test, sui2024just}, following MEMO \cite{zhang2022memo}, augment the test instance and aim to minimize marginal entropy, which is defined as the entropy of the mean prediction.
We decompose marginal entropy into two components: a pointwise entropy term and the Kullback–Leibler (KL) divergence, which measures the divergence between predictions from each augmented view and the mean prediction.
We observe that when adapting to adversarial examples with high-confidence inaccurate predictions, the KL divergence term tends to pull the augmented views toward the misleading mean prediction, which does not exist in the clean scenario.
This meaningless operation introduces conflicts into the optimization process.
To mitigate the influence of adversarial samples and preserve simplicity, we discard the KL divergence term, retaining only the pointwise entropy minimization for tuning textual prompts.
This straightforward modification not only defends against adversarial attacks but also maintains clean performance.

In order to effectively leverage knowledge from augmented views, we propose a reliability-based weighted ensembling strategy.
To assign a larger weight to reliable augmented views during ensembling, we introduce a similarity-based metric to assess the reliability of samples.
We hypothesize that samples with higher similarity to their neighbors are farther away from outliers and contain more reliable information.
Thus, we calculate the average similarity of each sample with its neighbors to estimate its reliability.
Using this metric, outliers such as adversarial examples and noisy augmented views are assigned lower reliability scores, which means less participation in the ensembling.
Finally, we obtain the final prediction by ensembling the individual model predictions, weighted according to their reliability.
Extensive experiments on fine-grained classification benchmarks and distribution shift benchmarks validate the effectiveness of our method in both adaptation and defense against adversarial attacks. Our contributions are summarized as follows:
\begin{itemize}
    \item We propose R-TPT, which is the first to explore test-time paradigms for defending against potential adversarial attacks in CLIP.
    \item We discard the KL divergence term from the marginal entropy objective to eliminate optimization conflicts and propose a reliability-based weighted ensembling strategy to integrate knowledge from reliable augmented views.
    \item Extensive experiments demonstrate the effectiveness of our method in both adaptation and adversarial defense.
\end{itemize}

\begin{figure*}[t]
    \centering
    \includegraphics[width=0.95\linewidth]{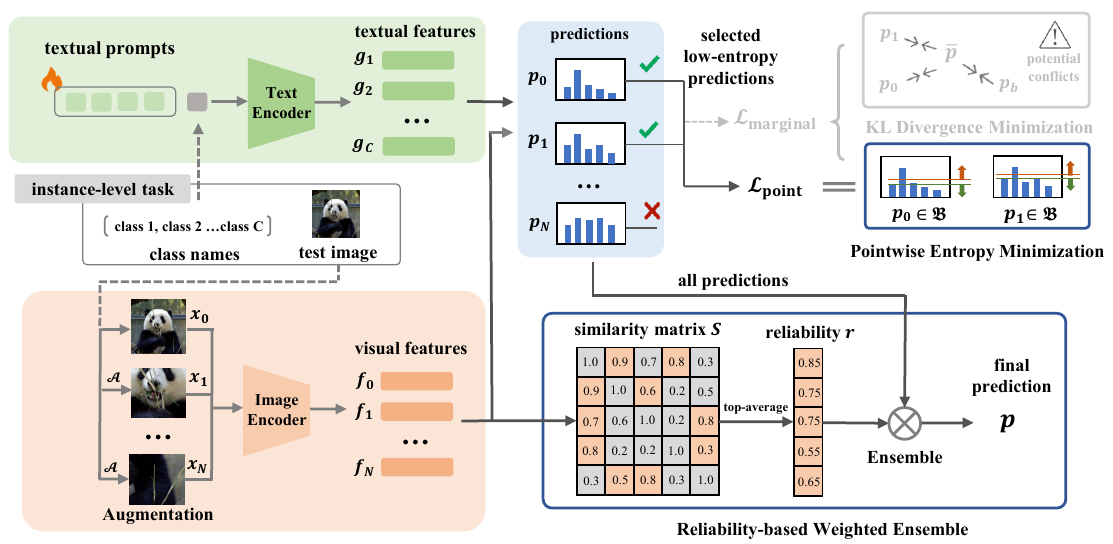}
    \caption{The pipeline of R-TPT framework. Given an instance-level task, we deploy augmentation on the test image and build a classifier with CLIP's text branch. After selecting a low-entropy batch, R-TPT discards the KL divergence minimization term which potentially introduces conflicts in the marginal entropy \cite{zhang2022memo} and optimizes textual prompts with pointwise entropy minimization. To effectively utilize the knowledge of the augmented views, R-TPT applies a reliability-based weighted ensembling mechanism in the final inference process.}
    \label{fig:method}
\end{figure*}

\section{Related Work}
\subsection{Adversarial Attack and Defense}
A lot of research \cite{szegedy2013intriguing, goodfellow2014explaining, madry2017towards} has been devoted to studying neural network's vulnerability to adversarial noise.
A pioneering work \cite{szegedy2013intriguing} introduces the concept of adversarial examples and finds small-amplitude noise that humans cannot recognize leads to misclassifications.
Since then, a series of attack methods to generate adversarial samples \cite{goodfellow2014explaining, madry2017towards, carlini2017towards, croce2020reliable} have been proposed.
FGSM \cite{goodfellow2014explaining} proposes to utilize the gradient sign to generate an adversarial example.
Researchers \cite{madry2017towards} generate the adversarial noise by projected gradient descent (PGD) operation, which becomes the standard measurement of model robustness.
Moreover, a variety of works explore adversarial attacks in many restricted conditions, ranging from one-pixel attack \cite{su2019one}, universal perturbation \cite{moosavi2017universal} to more realistic black-box setting \cite{andriushchenko2020square, ilyas2018black}.

In parallel, numerous defense strategies \cite{madry2017towards, zhang2019theoretically, wu2020adversarial, sheng2023adaptguard} have been proposed to mitigate adversarial attacks.
Adversarial training \cite{goodfellow2014explaining, madry2017towards} improves the robustness of the model by incorporating the adversarial samples into the training set.
TRADES \cite{zhang2019theoretically} provides a theoretical analysis of adversarial error to trade adversarial robustness against accuracy.
AWP \cite{wu2020adversarial} perturbs the model's weights with small adversarial noise during training to enhance robustness.
Reconstruction \cite{nie2022diffusion, samangouei2018defense} with generative models \cite{song2020score, goodfellow2014generative} is also a commonly used technique in test-time defense methods.

As for the vison-language models, TeCoA \cite{mao2022understanding} and APT \cite{li2024one} employ adversarial training to pretrained CLIP \cite{radford2021learning} to improve the adversarial robustness.
However, their training-time solution requires an annotated dataset and lacks flexibility across tasks.
In this work, we choose to deploy adversarial defense in the test time that is more effective and flexible and can collaborate with their training-time defense.

\subsection{Test-Time Adaptation for VLMs}
Test-time adaptation \cite{liang2024comprehensive, yu2023benchmarking, shu2022test} aims to adapt pre-trained models to the test data at inference time to improve the performance further.
According to the test data form, test-time adaptation is divided into streaming data adaptation \cite{wang2020tent, wang2022continual, yu2024stamp} and single instance adaptation \cite{zhang2022memo, shu2022test}, and our work focuses on the latter.
Recent research \cite{shu2022test, zanella2024test, sui2024just} has been devoted to exploring the instance adaptation methods for VLMs \cite{radford2021learning, wang2023large, zhang2024vision}.
TPT \cite{shu2022test} employs the marginal entropy minimization to augmentation views of the test instance to correct the prediction.
DiffTPT \cite{feng2023diverse} utilizes the diffusion technique to obtain diverse views which is helpful for the adaptation. 
PromptAlign \cite{abdul2024align} aligns the statistics of the test instance and collected natural images to make the model's parameters adapt to test samples.
To take advantage of more prompt templates \cite{radford2021learning}, TPS \cite{sui2024just} chooses to optimize feature shift and utilizes prompt ensembling for initialization.
Moreover, researchers \cite{zanella2024test} propose a training-free adaptation method by ensembling the augmented views with the MeanShift algorithm \cite{comaniciu1999mean}.

Besides accuracy, researchers \cite{yoon2024c} also focus on improving calibration performance through higher text feature dispersion.
In this work, we first utilize the test-time paradigm to defend against adversarial attacks for CLIP due to its severe vulnerability to adversarial examples.

\section{Methodology}

In this paper, we improve CLIP's adversarial robustness through a test-time paradigm, motivated by its inherent vulnerabilities and the high resource demands of train-time defense methods.
Our proposed robust test-time prompt tuning (R-TPT), as illustrated in Figure~\ref{fig:method}, requires no annotated data and is equipped with greater flexibility.
In Sec. 3.1, we begin with reviewing the foundational concepts of CLIP and the test-time prompt tuning approach.
We then introduce the two key components of R-TPT: pointwise entropy minimization in Sec. 3.2, and the reliability-based weighted ensembling strategy in Sec. 3.3.

\subsection{Preliminary}

\textbf{Contrastive language-image pre-training.}
CLIP is a popular language-vision model with a double-tower architecture, consisting of an image encoder $F(\cdot)$ and a text encoder $G(\cdot)$.
It is pretrained by the contrastive learning objective with a large amount of image-text pairs.
Benefiting from rich pretraining knowledge, CLIP has a strong zero-shot generalization ability.
Take a C-way classification task with class names $\{t_c\}_{c=1}^C$ as an example, CLIP obtains textual features $g_c$ by the text encoder $G(\cdot)$ with a prompt template (\eg, ``a photo of a []'') and the class name $t_c$ as the input.
Also, each test image $x_i$ queries the image encoder to calculate the image feature $f_i = F (x_i)$.
The probability that $x_i$ belongs to category $c$ is calculated by a softmax operation with the cosine similarity of those features:
\begin{equation} 
\label{eq:clipzs}
p_c(x_i) = \frac{\exp(cos(f_i, g_c) / \tau)} {\sum_{j=1}^{C} \exp(cos(f_i, g_j) / \tau)},
\end{equation}
where $cos(\cdot)$ represents the cosine similarity operation and $\tau$ refers to the temperature default set to 0.01.

\textbf{Test-time prompt tuning.}
Although CLIP exhibits strong classification performance, it is sensitive to distribution shifts.
To address this issue, test-time prompt tuning (TPT) \cite{shu2022test} improves the model's performance on individual test instances, without requiring additional training data for adaptation.
During the test time, TPT deploys augmentation operations on the test instance $x_0$ to obtain augmented views $\{x_i\}_{i=1}^N$ and tunes the textual prompts with low-entropy views.
The core objective of TPT is to minimize marginal entropy, which is formulated as:
\begin{equation}
\label{eq:mem}
\mathcal{L}_{marginal} = \mathcal{H}(\bar{p}) = \mathcal{H}(\frac{1}{\mid\mathcal{B}\mid} \sum_{x \in \mathcal{B}}{p(x)}),
\end{equation}
where $\mathcal{H}(\cdot)$ is the Shannon entropy and $\mathcal{B}$ represents the sample set selected from all views $\{x_i\}_{i=0}^N$ based on the low entropy.
When augmentation provides valuable information, the classification boundary adapts according to the new prompt, leading to more accurate predictions.

\subsection{Refining Marginal Entropy Minimization}
In the optimization of a single test sample, marginal entropy is the default choice for both visual models \cite{zhang2022memo} and multi-modal models \cite{shu2022test}.
Minimizing marginal entropy $\mathcal{H}(\bar{p})$ encourages the model to produce a consistent output across a set of selected low-entropy augmented views $\mathcal{B}$.
However, when the test sample is adversarially perturbed, it can easily be selected into the set $\mathcal{B}$.
While random augmentations can weaken the adversarial noise, enforcing consistency across all augmented outputs may mislead the optimization.
Since our goal is to utilize test time adaptation to strengthen the model's defense ability to adversarial examples, we propose refining the marginal entropy objective.

We decompose the marginal entropy objective into two items as follows:
\begin{equation}
\begin{aligned}
\label{eq:mem-decompose}
\mathcal{H}(\bar{p}) &= -\sum_{c=1}^{C} {\bar{p}_c} \ {log} {\bar{p}_c} = - \frac{1}{|\mathcal{B}|} \sum_{b=1}^{{|\mathcal{B}|}} \sum_{c=1}^{C} {p_c^b} \ {log} {\bar{p}_c}\\
&= \frac{1}{|\mathcal{B}|} \sum_{b=1}^{{|\mathcal{B}|}} \left( - \sum_{c=1}^{C} {p_c^b} \ {log} {p_c^b} \ +  \sum_{c=1}^{C} {p_c^b} \ {log}\frac{p_c^b}{\bar{p}_c} \right)  \\
&= \frac{1}{|\mathcal{B}|} \sum_{b=1}^{{|\mathcal{B}|}} \left( \mathcal{H}({p^b}) \ + \mathcal{KL} (p^b \| \bar{p})\right),
\end{aligned}
\end{equation}
where $\mathcal{H}(\cdot)$ denotes the Shannon entropy, $\mathcal{KL}(\cdot \| \cdot)$ refers to Kullback–Leibler (KL) divergence.
Eq.~\ref{eq:mem-decompose} shows that marginal entropy is a combination of a pointwise entropy term and a KL divergence term.
Minimizing pointwise entropy helps move the classification boundary away from low-entropy points, which is the main composition of the marginal entropy.
The KL divergence term, on the other hand, encourages consistent predictions across the augmented views.
In the case of clean test samples, the differences between low-entropy augmented views are small, and the KL divergence term has a minimal effect.
However, in adversarial scenarios, the mean prediction is distorted by the original perturbed sample, which differs significantly from the augmented views.
As a result, enforcing consistency across the augmented views leads to conflicts and neglects the valuable information in these views.

To improve performance under both clean and adversarial conditions, we discard the KL divergence term and focus solely on minimizing pointwise entropy, as follows:
\begin{equation}
\label{eq:ent}
minimize \ \mathcal{L}_{point} = \frac{1}{|\mathcal{B}|} \sum_{b=1}^{{|\mathcal{B}|}} \mathcal{H}({p^b}).
\end{equation}
For clean test instances, minimizing pointwise entropy functions similarly to the original marginal entropy objective.
Also, for adversarial examples, this approach focuses more on augmented views, which have relatively high entropy and information, and ignores the original adversarial inputs.
Therefore, our objective can handle both natural conditions and adversarial attacks well.

\begin{figure*}[!t]
		\small
		\setlength\tabcolsep{1mm}
		\renewcommand\arraystretch{0.1}
		\begin{tabular}{ccc}
            \includegraphics[width=0.32\linewidth]{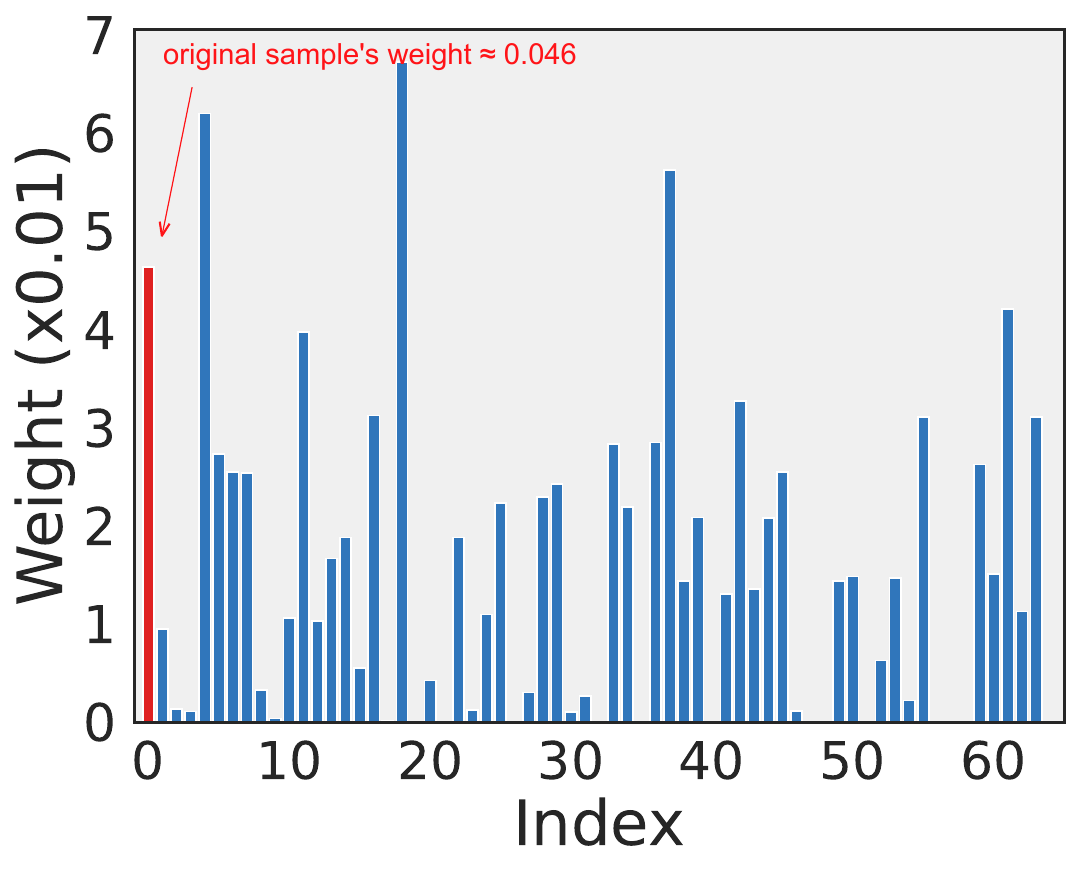} &		
            \includegraphics[width=0.32\linewidth]{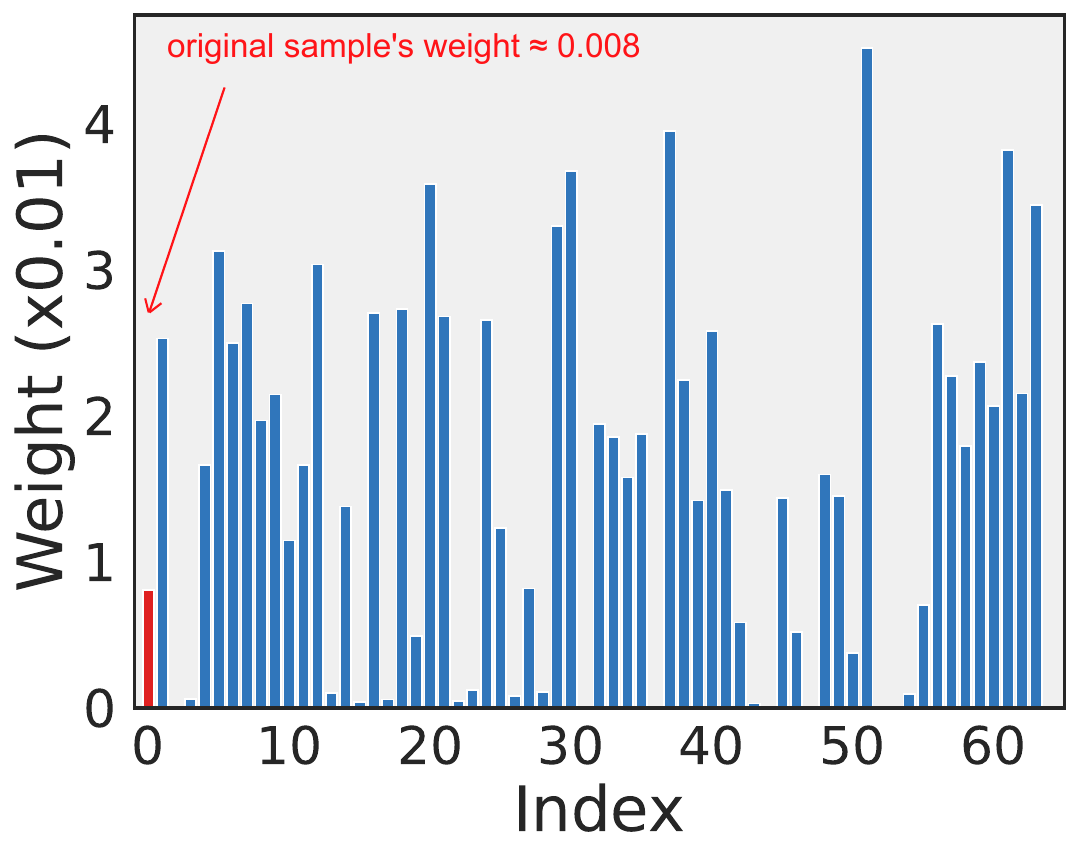} &
            \includegraphics[width=0.32\linewidth]{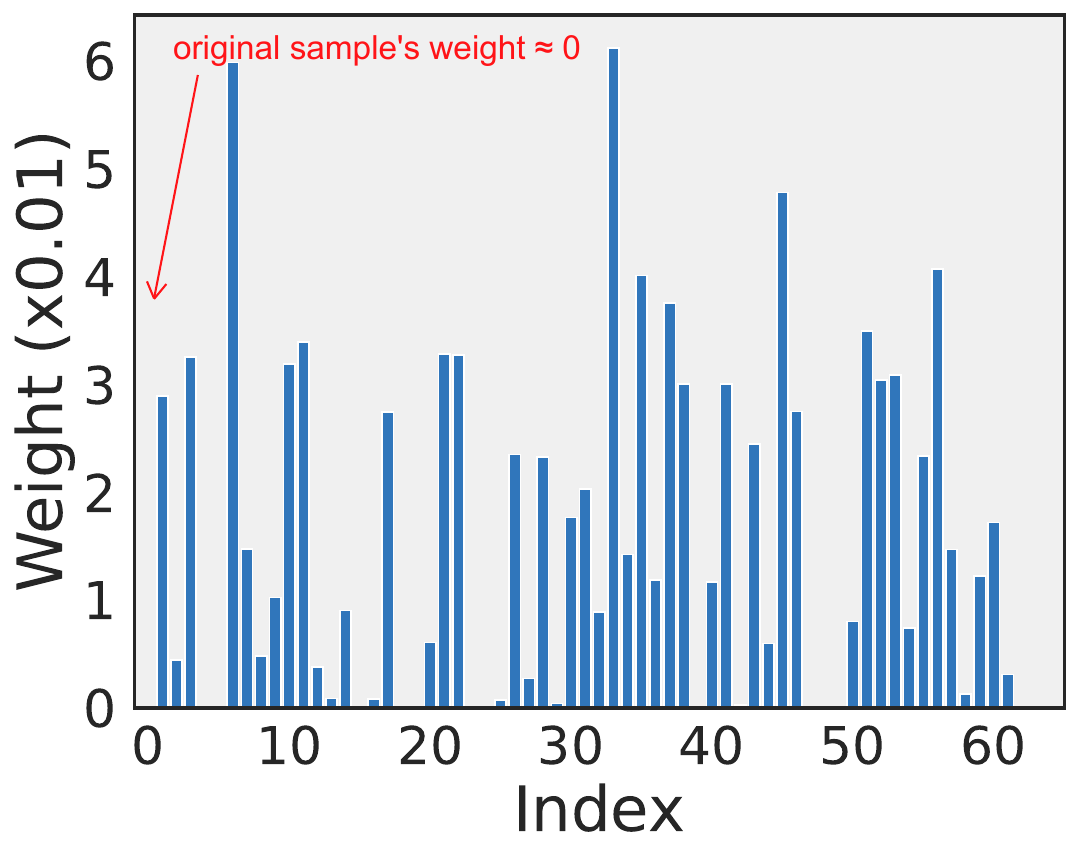}  
			\\
			(a) Weight value in ImageNet. & (b) Weight value in ImageNet-A. & (c) Weight value in ImageNet under attack.  \\
		\end{tabular}
		\caption{Visualization of assigned reliability-based weight value under (a) clean, (b) distribution shift, and (c) adversarial attack scenarios. The red and blue bars denote the weight value of the original test instance and its augmented views, respectively. The weight assigned to the clean instance is higher than under the distribution shift. The weight of the adversarial test sample is close to 0.}
		\label{fig:weight}
\end{figure*}

\subsection{Reliability-based Weighted Ensembling}
Unsupervised prompt tuning typically makes no change to image features and a small adjustment to the classification boundary.
For adversarial examples, it is challenging to correct incorrect predictions since their feature representations are far from the correct decision boundary.
Fortunately, augmentation at the pixel level can help mitigate the adversarial noise, as it weakens the effect of perturbations.
Thus, we can leverage diverse augmented views of the test image, which provide valuable knowledge.
To integrate information from augmented views effectively and protect the ensembling from the noise of lower-quality views, we propose a reliability-based weighted ensembling mechanism.

Since augmentation inherently involves randomness (e.g., background areas left after cropping), we introduce a reliability metric to represent the quality of each augmented view.
Reliability is defined as the similarity between the sample and its nearest neighbors.
High reliability indicates that the sample is densely clustered in the feature space, containing useful information.
Conversely, low reliability suggests that the sample is an outlier, likely containing augmentation or adversarial noise, and should be down-weighted or ignored during ensembling.

Given an augmented batch, which includes the test image and its $N$ augmented views, we calculate the similarity matrix $\{S_{i,j}\}_{i,j=0}^{N}$ of the visual features $\{f_i\}_{i=0}^N$ :
\begin{equation}
{S_{i,j}} = cos( f_i, f_j ), \ \ i,j = 0,1,...,N \ ,
\end{equation}
where $con(\cdot, \cdot)$ denotes the cosine similarity operation.
The $i$-th row of the matrix $S$ represents the similarity between $x_i$ and all samples in the batch.
We then select the $K$ closest neighbor of $x_i$ (excluding $x_i$ itself) to form the neighboring set $\mathcal{N}_i$.
The reliability of $x_i$ is calculated as the average similarity within the neighboring set $\mathcal{N}_i$:
\begin{equation}
{r_{i}} = \frac{1}{K} \sum_{k \in \mathcal{N}_i} cos( f_i, f_k ), \ \ i = 0,1,...,N \ .
\end{equation}
The reliability score reflects the degree to which $x_i$ is surrounded by similar samples in the feature space.

To obtain the final prediction, we ensemble the predictions from all augmented views, weighted by their respective reliability scores.
The weight assigned to each view is calculated by applying a softmax operation on the reliabilities $\{r_i\}_{i=0}^N$.
This weight assignment mechanism is flexible and adapts well to various scenarios.
We present the weight values for clean samples (ImageNet), samples with distribution shift (ImageNet-A), and adversarial samples (ImageNet) in Figure~\ref{fig:weight}.
As shown, clean samples are assigned higher weights, and the ensembling mechanism focuses on augmented views when the test instance shows significant distribution shifts from natural images.
For adversarial samples, the weight assigned to the poisoned instance is close to zero, which indicates that our mechanism protects the ensembling process from being misled by adversarial noise.
This approach not only improves the robustness of predictions on adversarial samples but also maintains strong performance on clean samples, thus enhancing the overall reliability of the inference process.

\setlength{\tabcolsep}{4.0pt}
\begin{table*}[ht]
\vspace{-5pt}
  \centering
  \resizebox{0.98\linewidth}{!}{%
  \begin{tabular}{l|ca|ca|ca|ca|ca|ca|ca|ca|ca}
    \toprule
    \multirow{2}{*}{Method} & \multicolumn{2}{c|}{Caltech101} & \multicolumn{2}{c|}{Pets} & \multicolumn{2}{c|}{Cars} & \multicolumn{2}{c|}{Flower102} & \multicolumn{2}{c|}{Aircraft} & \multicolumn{2}{c|}{DTD} & \multicolumn{2}{c|}{EuroSAT} & \multicolumn{2}{c|}{UCF101} & \multicolumn{2}{c}{Avg.} \\
    & Acc. & \multicolumn{1}{c|}{Rob.} & Acc. & \multicolumn{1}{c|}{Rob.} & Acc. & \multicolumn{1}{c|}{Rob.} & Acc. & \multicolumn{1}{c|}{Rob.} & Acc. & \multicolumn{1}{c|}{Rob.} & Acc. & \multicolumn{1}{c|}{Rob.} & Acc. & \multicolumn{1}{c|}{Rob.} & Acc. & \multicolumn{1}{c|}{Rob.} & Acc. & \multicolumn{1}{c}{Rob.} \\
    \midrule

    CLIP \cite{radford2021learning} & 85.9 & 2.6  & 83.5 & 0.0  & 55.7 & 0.0  & 61.7 & 0.0  & 15.7 & 0.0 & 40.4 & 0.8  & 23.7 & 0.0  & 58.9 & 0.0  & 53.2 & 0.4  \\
    Ensemble & 83.5 & 74.8 & 82.3 & 69.9 & 57.1 & 36.2 & 58.0 & 46.6 & 16.4 & 9.8 & 37.1 & 29.5 & 16.7 & 13.7 & 53.9 & 43.0 & 50.6 & 40.4 \\
    TPT \cite{shu2022test} & \bestclean{87.9} & 7.0  & 84.7 & 0.1  & 58.4 & 0.0  & 62.1 & 0.0  & 17.3 & 0.0 & \bestclean{42.4} & 4.3  & \bestclean{28.4} & 0.0  & \bestclean{60.6} & 0.3  & 55.2 & 1.5  \\
    C-TPT \cite{yoon2024c} & 87.7 & 3.7  & 83.6 & 0.0  & 56.6 & 0.0  & \bestclean{64.8} & 0.0  & 16.7 & 0.0 & 41.5 & 1.3  & 27.0 & 0.0  & 60.1 & 0.1  & \bestclean{54.8} & 0.6  \\
    MTA \cite{zanella2024test} & 87.3 & 65.9 & \bestclean{84.8} & 59.8 & \bestclean{58.7} & 17.8 & 61.0 & 31.5 & \bestclean{18.1} & 3.7 & 40.3 & 18.8 & 22.5 & 1.6  & \bestclean{60.6} & 31.3 & 54.1 & 28.8 \\
    R-TPT & 86.7 & \best{79.8} & 84.6 & \best{74.2} & 58.1 & \best{42.9} & 60.6 & \best{51.9} & 17.5 & \best{12.6} & 41.3 & \best{33.5} & 21.2 & \best{15.9} & 59.7 & \best{50.9} & 53.7 & \best{45.2} \\
  
    \bottomrule
  \end{tabular}
}
\caption{Results (\%) of clean accuracy (Acc.) and adversarial accuracy (Rob.) of various adaptation methods on \textbf{fine-grained classification datasets} with pre-trained CLIP-ResNet50 ($\epsilon$ = 1.0). Best clean accuracies are (\bestclean{bold}), best adversarial accuracies are (\best{bold red}).}
\label{tab:fine-grained}
\end{table*}

\setlength{\tabcolsep}{4.0pt}
\begin{table*}[ht]
\vspace{-5pt}
  \centering
  \resizebox{0.98\linewidth}{!}{%
  \begin{tabular}{l|ca|ca|ca|ca|ca|ca|ca|ca|ca}
    \toprule
    \multirow{2}{*}{Method} & \multicolumn{2}{c|}{Caltech101} & \multicolumn{2}{c|}{Pets} & \multicolumn{2}{c|}{Cars} & \multicolumn{2}{c|}{Flower102} & \multicolumn{2}{c|}{Aircraft} & \multicolumn{2}{c|}{DTD} & \multicolumn{2}{c|}{EuroSAT} & \multicolumn{2}{c|}{UCF101} & \multicolumn{2}{c}{Avg.} \\
    & Acc. & \multicolumn{1}{c|}{Rob.} & Acc. & \multicolumn{1}{c|}{Rob.} & Acc. & \multicolumn{1}{c|}{Rob.} & Acc. & \multicolumn{1}{c|}{Rob.} & Acc. & \multicolumn{1}{c|}{Rob.} & Acc. & \multicolumn{1}{c|}{Rob.} & Acc. & \multicolumn{1}{c|}{Rob.} & Acc. & \multicolumn{1}{c|}{Rob.} & Acc. & \multicolumn{1}{c}{Rob.} \\
    \midrule
    CLIP \cite{radford2021learning} & 94.0 & 0.0  & \bestclean{88.3} & 0.0  & 65.5 & 0.0  & 67.4 & 0.0  & 23.9 & 0.0 & 44.4 & 0.0  & 42.2 & 0.0 & 65.2 & 0.0  & 61.4 & 0.0 \\
    Ensemble & 91.9 & 74.7 & 86.2 & 51.2 & 65.7 & 26.0 & 65.9 & 36.3 & 23.4 & 8.7 & 43.2 & 25.1 & 28.2 & 2.2 & 63.0 & 30.6 & 58.4 & 31.8 \\
    TPT \cite{shu2022test} & 94.1 & 0.0  & 87.4 & 0.0  & 66.5 & 0.0  & 69.1 & 0.0  & 23.4 & 0.0 & \bestclean{46.9} & 0.0  & \bestclean{42.6} & 0.0 & \bestclean{67.9} & 0.0  & 62.2 & 0.0  \\
    C-TPT \cite{yoon2024c} & 93.9 & 0.0  & 88.2 & 0.0  & 65.8 & 0.0  & \bestclean{69.6} & 0.0  & 23.9 & 0.0 & 45.9 & 0.0  & 42.3 & 0.0 & 65.5 & 0.0  & 61.9 & 0.0  \\
    MTA \cite{zanella2024test} & \bestclean{94.3} & 72.1 & 88.0 & 51.8 & \bestclean{67.7} & 18.5 & 67.4 & 27.9 & \bestclean{25.0} & 4.3 & 46.5 & 16.2 & 42.5 & 1.2 & 67.5 & 27.5 & \bestclean{62.3} & 27.4 \\
    R-TPT  & 93.7 & \best{82.0} & 87.2 & \best{60.2} & 67.0 & \best{34.7} & 68.7 & \best{44.6} & 23.9 & \best{13.2} & 46.4 & \best{32.8} & 34.7 & \best{8.5} & 67.2 & \best{43.2} & 61.1 & \best{39.9} \\
    \bottomrule
  \end{tabular}
}
\caption{Results (\%) of various adaptation methods on \textbf{fine-grained classification datasets} with pre-trained CLIP-ViT-B/16 ($\epsilon$ = 4.0).}
\label{tab:vitb16}
\end{table*}

\section{Experiment}

\subsection{Setup}

\textbf{Datasets.}
To evaluate our proposed test-time defense method for VLMs, we conduct experiments on eight fine-grained classification datasets.
These databases cover general objects (\textbf{Caltech101}~\cite{fei2004learning}), animals (\textbf{Pets}~\cite{parkhi2012cats}), plants (\textbf{Flower102}~\cite{nilsback2008automated}), vehicles (\textbf{Cars}~\cite{krause20133d}, \textbf{Aircraft}~\cite{maji2013fine}), textures (\textbf{DTD}~\cite{cimpoi2014describing}), satellite images (\textbf{EuroSAT}~\cite{helber2019eurosat}), and actions from videos (\textbf{UCF101}~\cite{soomro2012ucf101}).
Moreover, we evaluate on \textbf{ImageNet} \cite{deng2009imagenet} and four ImageNet-out-of-distribution (OOD) benchmarks with distribution shift.
\textbf{ImageNet-A} \cite{hendrycks2021natural} contains 200 classes and 7,500 natural adversarial examples which are collected with adversarial filtration technique.
\textbf{ImageNet-V2} \cite{recht2019imagenet} consists of 10,000 natural images across 1,000 categories from a different source than ImageNet.
\textbf{ImageNet-R} \cite{hendrycks2021many} is a dataset containing 30,000 images with various renditions, leading to different textures and local statistics from ImageNet.
\textbf{ImageNet-S} \cite{wang2019learning} consists of 50,889 sketch-style images and shares the same category space with ImageNet.
Since our method is to defend against potential adversarial attacks during the test time, we do not need access to the training set of the above datasets.

\textbf{Evaluation metrics.}
Since our task focuses on instance-level test-time adaptation, the model update and prediction of each test sample can not utilize the information of other samples.
Following the previous works \cite{shu2022test, zanella2024test}, we report the average classification accuracy (\textbf{Acc.}) to measure the method's adaptation ability on clean samples.
To evaluate the adversarial defense performance, we provide the average accuracy (\textbf{Rob.}) on adversarial examples generated by the PGD algorithm \cite{madry2017towards} with various noise radii.
It is worth noting that the adversarial examples are calculated on CLIP before adaptation, which is more suitable for real-world applications since attackers rely on open-source models and have no idea about the victim model's algorithm.

\textbf{Baselines.}
We compare R-TPT on the above benchmarks with existing test-time adaptation methods for CLIP, including TPT~\cite{shu2022test}, C-TPT~\cite{yoon2024c}, and MTA~\cite{zanella2024test}, as well as zero-shot prediction from CLIP~\cite{radford2021learning}.
Also, we regard Ensemble as an additional baseline method, which employs simple average operation on predictions of all augmented views.
Note that both our method and the compared baseline methods rely only on CLIP and AugMix \cite{hendrycks2019augmix} augmentation, without any additional foundation models (\eg, LLM, diffusion models) or knowledge.
All results of baselines are reproduced with the official code.

\textbf{Implementation details.}
For all experiments, we utilize official pre-trained CLIP-ResNet50 and CLIP-ViT-B/16 \cite{radford2021learning} as our base model.
As for adversarial example generation, we utilize the PGD algorithm \cite{madry2017towards} with $\epsilon$ = 1.0 and 7 steps for ResNet, while $\epsilon$ = 4.0 and 100 steps for ViT.
In the defense stage at test time, the parameter optimized in all experiments is a textual prompt with a context length of 4 and is initiated with ``a photo of a''.
We adopt the Adam optimizer with weight decay and a single step.
Following previous work \cite{shu2022test}, the learning rate is set to 0.005 and the augmented batch size is 64.
All experiments use the PyTorch framework and run on RTX3090 GPUs.

\setlength{\tabcolsep}{6.0pt}
\begin{table*}[ht]
\vspace{-5pt}
  \centering
  \resizebox{0.8\linewidth}{!}{%
  \begin{tabular}{l|ca|ca|ca|ca|ca|ca}
    \toprule
    \multirow{2}{*}{Method} & \multicolumn{2}{c|}{ImageNet} & \multicolumn{2}{c|}{ImageNet-A} & \multicolumn{2}{c|}{ImageNet-V2} & \multicolumn{2}{c|}{ImageNet-R} & \multicolumn{2}{c|}{ImageNet-S} & \multicolumn{2}{c}{Avg.} \\
    & Acc. & \multicolumn{1}{c|}{Rob.} & Acc. & \multicolumn{1}{c|}{Rob.} & Acc. & \multicolumn{1}{c|}{Rob.} & Acc. & \multicolumn{1}{c|}{Rob.} & Acc. & \multicolumn{1}{c|}{Rob.} & Acc. & \multicolumn{1}{c}{Rob.} \\    
    \midrule
    CLIP \cite{radford2021learning} & 58.2 & 0.1  & 21.8 & 0.0  & 51.5 & 0.1  & 56.1 & 0.8  & 33.3 & 0.5  & 44.2 & 0.3  \\
    Ensemble & 58.0 & 40.1 & 22.6 & 10.1 & 52.0 & 37.2 & 51.3 & 39.3 & 29.5 & 20.7 & 42.7 & 29.5 \\
    TPT \cite{shu2022test} & 60.7 & 0.3  & 26.5 & 0.0  & 54.8 & 0.3  & \bestclean{58.9} & 1.8  & 35.0 & 1.4  & \bestclean{47.2} & 0.7  \\
    C-TPT \cite{yoon2024c} & 60.4 & 0.1  & 24.1 & 0.0  & 54.3 & 0.1  & 57.7 & 1.0  & 34.7 & 0.9  & 46.2 & 0.4  \\
    MTA \cite{zanella2024test} & 60.4 & 30.0 & 27.5 & 5.6  & 54.2 & 24.6 & 58.4 & 29.8 & \bestclean{35.2} & 11.3 & 47.1 & 20.3 \\
    R-TPT & \bestclean{60.9} & \best{47.7} & \bestclean{28.4} & \best{14.4} & \bestclean{54.9} & \best{41.6} & 57.6 & \best{46.9} & 34.0 & \best{26.2} & 47.1 & \best{35.4} \\
    \bottomrule
  \end{tabular}
}
\caption{Results (\%) of various adaptation methods on \textbf{ImageNet and ImageNet-OOD benchmarks} with pre-trained CLIP-ResNet50. OOD Avg. refers to the average results among four ImageNet-OOD benchmarks ($\epsilon$ = 1.0).}
\label{tab:imagetnet}
\end{table*}

\setlength{\tabcolsep}{4.0pt}
\begin{table*}[ht]
\vspace{-5pt}
  \centering
  \resizebox{\linewidth}{!}{%
  \begin{tabular}{l|ca|ca|ca|ca|ca|ca|ca|ca|ca}
    \toprule
    \multirow{2}{*}{Method} & \multicolumn{2}{c|}{Caltech101} & \multicolumn{2}{c|}{Pets} & \multicolumn{2}{c|}{Cars} & \multicolumn{2}{c|}{Flower102} & \multicolumn{2}{c|}{Aircraft} & \multicolumn{2}{c|}{DTD} & \multicolumn{2}{c|}{EuroSAT} & \multicolumn{2}{c|}{UCF101} & \multicolumn{2}{c}{Avg.} \\
    & Acc. & \multicolumn{1}{c|}{Rob.} & Acc. & \multicolumn{1}{c|}{Rob.} & Acc. & \multicolumn{1}{c|}{Rob.} & Acc. & \multicolumn{1}{c|}{Rob.} & Acc. & \multicolumn{1}{c|}{Rob.} & Acc. & \multicolumn{1}{c|}{Rob.} & Acc. & \multicolumn{1}{c|}{Rob.} & Acc. & \multicolumn{1}{c|}{Rob.} & Acc. & \multicolumn{1}{c}{Rob.} \\
    \midrule
    CLIP-TeCoA \cite{mao2022understanding} & 79.3 & 44.3 & \bestclean{66.9} & 15.8 & 10.2 & 1.0 & \bestclean{30.8} & 9.0  & 6.6 & 0.5 & 24.5 & 10.7 & \bestclean{14.5} & 10.8 & 34.6 & 6.7  & 33.4 & 12.3 \\
    Ensemble & 72.7 & 55.1 & 59.9 & 38.9 & 5.6  & 2.7 & 26.6 & 16.0 & 4.2 & 2.0 & 23.5 & 16.2 & 12.5 & 11.0 & 26.4 & 14.0 & 28.9 & 19.5 \\
    TPT \cite{shu2022test} & 79.3 & 52.7 & 65.2 & 27.4 & 9.6  & 2.0 & 27.9 & 12.3 & \bestclean{6.7} & 1.7 & 25.5 & 14.6 & 12.2 & 11.2 & 34.9 & 10.2 & 32.7 & 16.5 \\
    C-TPT \cite{yoon2024c} & \bestclean{79.8} & 47.3 & 66.1 & 19.5 & \bestclean{10.6} & 1.3 & 29.4 & 10.7 & 6.4 & 0.7 & \bestclean{26.2} & 12.4 & 13.0 & 11.1 & \bestclean{36.4} & 8.1  & \bestclean{33.5} & 13.9 \\
    MTA \cite{zanella2024test} & 79.7 & 55.7 & 66.2 & 31.2 & 9.0  & 2.5 & 29.1 & 14.0 & 6.5 & 1.6 & 24.4 & 13.5 & 13.3 & 11.2 & 34.6 & 12.5 & 32.9 & 17.8 \\
    R-TPT  & 76.1 & \best{60.5} & 63.2 & \best{40.1} & 7.7  & \best{3.5} & 26.6 & \best{16.5} & 6.1 & \best{2.7} & 25.2 & \best{17.7} & 11.5 & \best{11.3} & 31.1 & \best{17.4} & 30.9 & \best{21.2} \\
    \bottomrule
  \end{tabular}
}
\caption{Results (\%) of adaptation methods on \textbf{fine-grained classification datasets} with TeCoA pre-trained CLIP-ViT-B/32 ($\epsilon$ = 4.0).}
\label{tab:tecoa-vitb32}
\end{table*}

\subsection{Experimental Results}

\textbf{Results on fine-grained datasets.}
We evaluate the adaptation and adversarial defense ability of R-TPT on eight fine-grained benchmarks and report the results in Table~\ref{tab:fine-grained}.
It is shown that CLIP with strong zero-shot generalization ability is suffering from the adversarial attack with a small radius.
TPT can steadily improve the accuracy of clean images but has weak defense capability.
We observe that methods with an ensembling strategy (\eg, Ensemble, MTA, R-TPT) achieve significantly better adversarial accuracy.
In particular, R-TPT achieves a 45.8\% of adversarial accuracy which outperforms all baseline methods.
Besides the attractive defense performance, R-TPT also improves the clean accuracy of CLIP from 53.2\% to 53.7\%.
In contrast, although Ensemble has strong defense capabilities, it suffers from negative transfer in the clean scenario.

\textbf{Results on ImageNet and ImageNet-OOD datasets.}
We provide the experiential results on ImageNet and four ImageNet-OOD datasets in Table~\ref{tab:imagetnet}.
CLIP can overcome distribution shifts, but fails to defend against adversarial attacks.
R-TPT performs the best in adversarial defense under both ImageNet and its related out-of-distribution benchmarks.
Especially, R-TPT achieves an adversarial accuracy of 47.7\% on ImageNet, 7.6\% higher than the second-best method Ensemble, while CLIP's adversarial accuracy on this benchmark is 0.1\%.
Moreover, our method obtains a similar clean accuracy with TPT, indicating that R-TPT is also effective for distribution shifts.

\textbf{Results of CLIP-ViT backbone.}
We evaluate our method on CLIP-ViT-B/16 \cite{radford2021learning} model and provide the results in Table~\ref{tab:vitb16}.
We observe that R-TPT outperforms all baseline methods regarding adversarial defense performance.
However, the performance gain of all adaptation methods for clean samples is small, and R-TPT and the remaining two ensembling-based methods make weak negative transfers.
This illustrates that the space for improvement of clean accuracy for strong pre-training models is small, and R-TPT can greatly enhance its weak robustness.

\subsection{More Analysis}

\textbf{Results under robust-pretrained models.}
we report the results of deploying adaptation methods on CLIP-ViT-B/32 with TeCoA \cite{mao2022understanding} robust pretrained models in Table~\ref{tab:tecoa-vitb32}.
It is shown that TeCoA significantly improves the robustness of CLIP, and the defense effect against adversarial attacks can benefit from our method during the testing stage, which increases adversarial accuracy form 13.1\% to 21.7\%.
At the same time, we also find that R-TPT's defense effect on the TeCoA pretrained model is weaker than the vanilla CLIP.
The reason is that introducing adversarial learning during the training time decreases the clean accuracy, which is the upper bound of defense.
Robust pre-trained models also generate more difficult adversarial examples.

\textbf{Analysis under various attacks.}
To demonstrate the versatility, we investigate the defense performance of R-TPT and baseline methods under various attack methods.
We employ CW \cite{carlini2017towards}, DeepFool \cite{moosavi2016deepfool}, and FGSM \cite{goodfellow2014explaining} as new attacks and report the adversarial accuracy on three benchmarks in Table~\ref{tab:other-attack}.
It is shown that our method improves CLIP's defense capability under each attack, which is consistent with the trend of PGD.
In particular, R-TPT has reached 51.8\% adversarial accuracy on Flowers.
The excellent defense performance among various attack methods proves the versatility of R-TPT.

\begin{figure*}[!t]
		\small
		\setlength\tabcolsep{1mm}
		\renewcommand\arraystretch{0.1}
		\begin{tabular}{ccc}
            \includegraphics[width=0.32\linewidth]{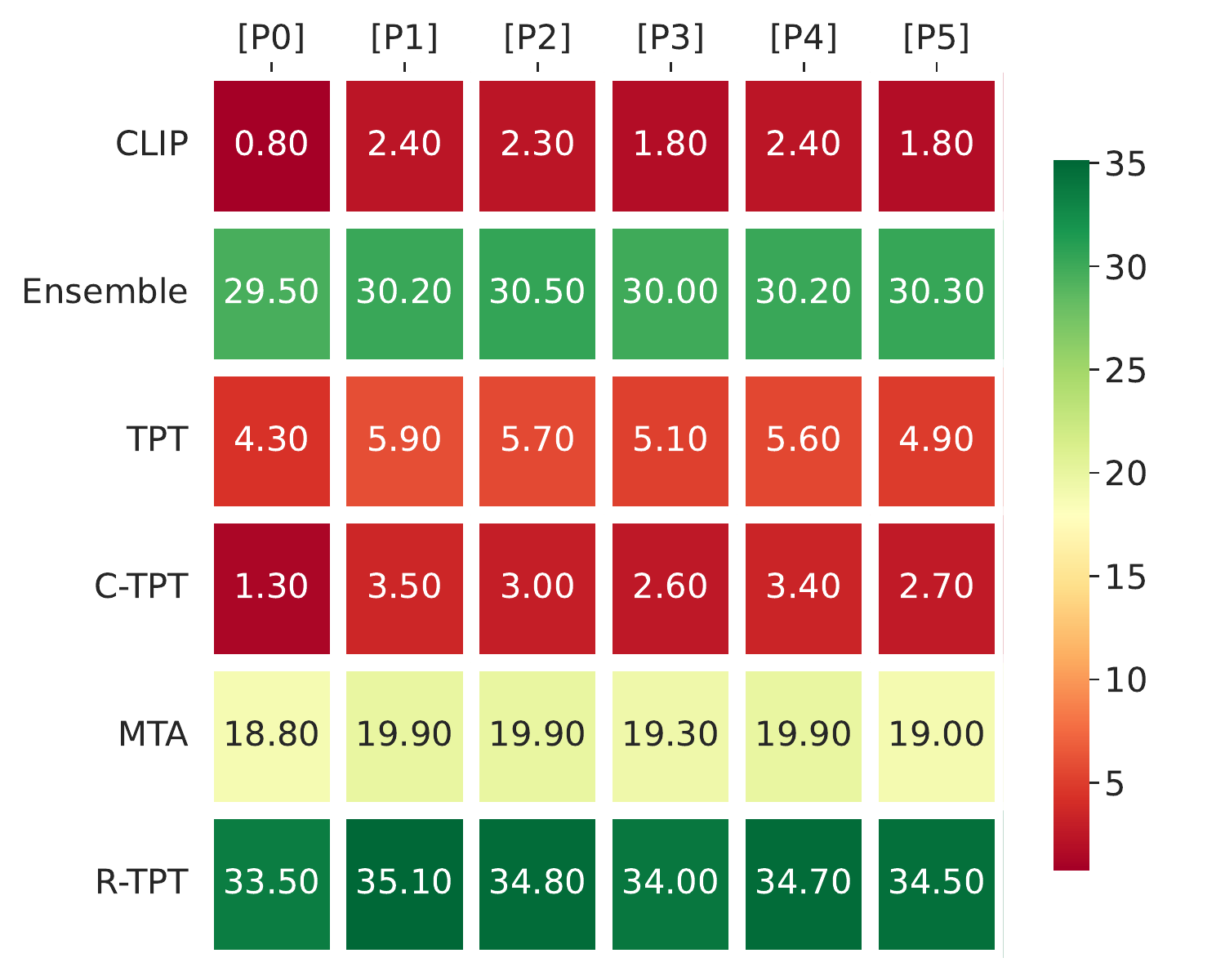} &		
            \includegraphics[width=0.32\linewidth, clip]{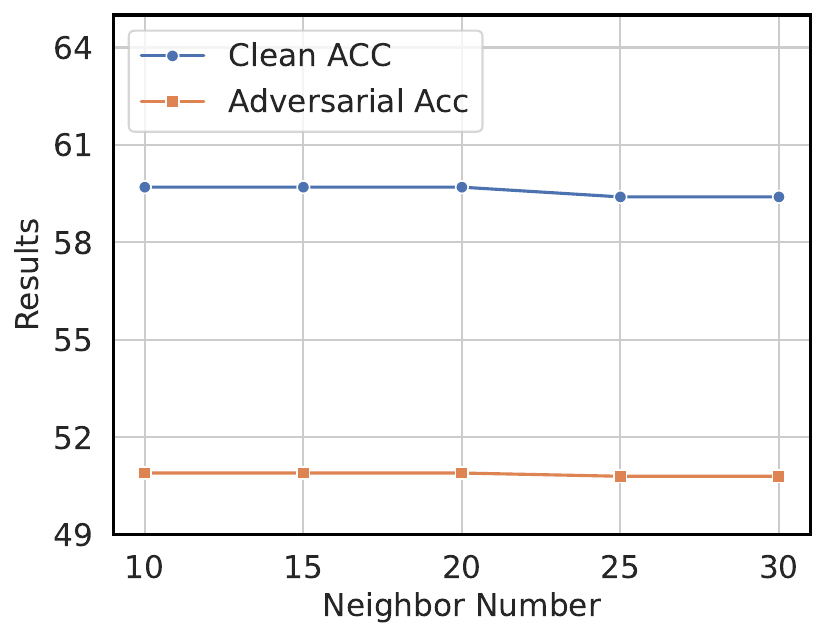} &
            \includegraphics[width=0.32\linewidth, clip]{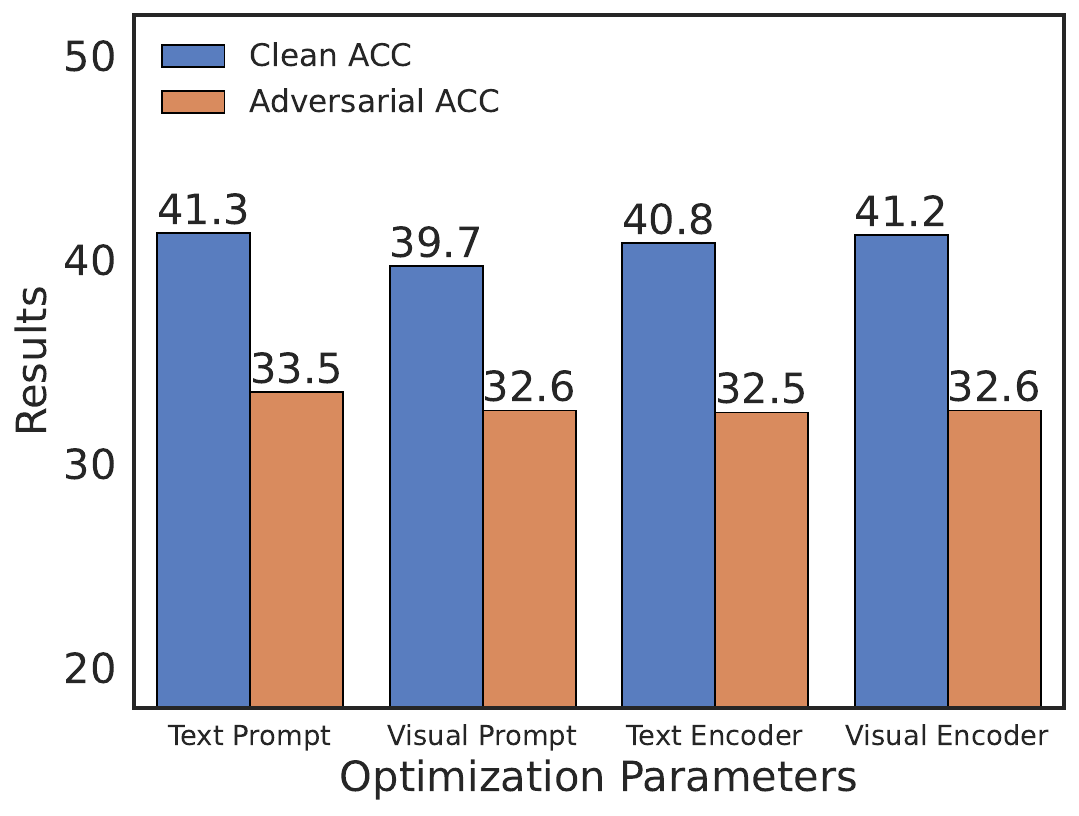}  
			\\[0.5mm]
			(a) Different prompt templates. & (b) Sensitivity of neighbor number. & (c) Sensitivity of optimization parameters.  \\
		\end{tabular}
		\caption{Results of different prompt templates in adversarial examples generation, different neighbor numbers, and different choices of optimization parameters in DTD dataset (CLIP-ResNet50, $\epsilon$ = 1.0).}
		\label{fig:exp}
\end{figure*}

\setlength{\tabcolsep}{3.0pt}
\begin{table}[ht]
\vspace{-5pt}
  \centering
  \resizebox{\linewidth}{!}{%
  \begin{tabular}{l|cccc|cccc}
    \toprule
        \multirow{2}{*}{Method} & \multicolumn{4}{c|}{Flowers} & \multicolumn{4}{c}{DTD} \\
         & CW & DF & FGSM & Avg. & CW & DF & FGSM & Avg. \\
    \midrule
        CLIP \cite{radford2021learning} & 0.8  & 0.4  & 4.8  & 2.0  & 2.3  & 7.6  & 13.4 & 7.8 \\
        Ensemble & 50.1 & 52.2 & 46.6 & 49.7 & 31.1 & 32.9 & 29.7 & 31.2 \\
        TPT \cite{shu2022test} & 13.8 & 10.8 & 14.2 & 12.9 & 21.3 & 24.4 & 22.2 & 22.6 \\
        C-TPT \cite{yoon2024c} & 6.6  & 5.5  & 6.2  & 6.1  & 11.9 & 15.8 & 17.5 & 15.1 \\
        MTA \cite{zanella2024test} & 34.5 & 35.4 & 36.6 & 35.5 & 23.6 & 23.5 & 23.9 & 23.7 \\
        R-TPT & \best{51.6} & \best{54.7} & \best{49.2} & \best{51.8} & \best{34.2} & \best{35.9} & \best{32.5} & \best{34.2} \\
    \bottomrule
  \end{tabular}
}
\caption{Adversarial accuracies (\%) of adaptation methods against different attacks on two fine-grained datasets (DF = DeepFool).}
\label{tab:other-attack}
\end{table}

\textbf{Analysis of inference efficiency.}
We study the inference efficiency of R-TPT.
Experiments of training-time defense method APT and R-TPT on UCF101 are provided in Table~\ref{tab:time}.
It is shown that R-TPT spends a certain amount of time on each test sample, which is different from training-time defense which spends a lot of resources at one time during training.
Moreover, we find that inference can be accelerated by appropriately reducing the number of augmented views to obtain better real-time performance.
When we choose to reduce the number of views from 64 to 16, the time required for each test sample is reduced to less than half of the original.

\setlength{\tabcolsep}{3.0pt}
    \begin{table}[!t]
        \centering
        \resizebox{\linewidth}{!}
        {
            \begin{tabular}{lc|cc}
            \toprule
            Method & Stage & Running time & Rob.\\
            \midrule
            APT+TeCoA (4shots) & Training time & 208s/50epochs & 39.4 \\
            R-TPT (64 views) & Test time & 0.58s/image & \color{blush}{41.0} \\
            R-TPT (32 views) & Test time & 0.28s/image & 40.8 \\
            R-TPT (16 views) & Test time & 0.20s/image & 40.0 \\
            \bottomrule
            \end{tabular}
        }
    \caption{Running time and adversarial accuracies (\%) of adaptation methods against adversarial attack on UCF101 dataset.}
    \label{tab:time}
    \end{table}

\textbf{Analysis under different prompt templates.}
Since in the realistic scenario, the attacker can not access the prompts of the model textual branch, we evaluate the defense performance of the adversarial examples generated by different prompt templates.
The result of DTD benchmark with six textual templates is provided in Figure~\ref{fig:exp}(a) (P0=`a photo of a []', P1=`a bad photo of the []', P2=`a origami []', P3=`a photo of the large []', P4=`a toy []', P5=`art of the []').
It is shown that CLIP is suffering from adversarial samples even if attackers use different prompt templates.
Also, R-TPT outperforms all baseline methods under each template.

\textbf{Analysis of sensitivity of hyperparameters and optimization parameters.}
We study the impact of the neighbor number in calculating reliability and the optimization parameter to further demonstrate the effectiveness of R-TPT.
We provide the results of R-TPT with different neighbor numbers (10, 15, 20, 25, 30) in Figure~\ref{fig:exp}(b).
The results demonstrate that R-TPT performs stable across different neighbor numbers in terms of both clean and adversarial accuracy.
Please note that if the neighbor number equals to augmented batch size, it will assign all samples with equal weights and the reliability-based ensembling mechanism will degenerate into the vanilla ensembling.
Besides, we report the results with different optimization parameters in Figure~\ref{fig:exp}(c).
Compared to other choices of the parameter space, optimizing textual prompts owns better defense and adaptation effects and fewer parameters, which make it always a popular solution for parameter-efficient fine-tuning.

\subsection{Ablation Study}

To study the contribution of terms in our method, we investigate the effectiveness of ensembling, reliability-based weighted mechanism, and pointwise entropy objective in R-TPT.
The results of the ablation study are reported in Table~\ref{tab:ablation}.
From the table, we find that entropy minimization focuses on enhancing the accuracy of adaptation on the clean samples and Slightly improves the defense performance on CLIP.
However, When cooperating with weighted ensembling, the improvement of entropy minimization on both metrics is significant.
The ensembling strategy greatly strengthens the defense capabilities of the model, but it also reduces the adaptation performance on clean samples.
The weighted mechanism gives reliable samples more important roles during prediction, further improving the defense capabilities and mitigating the clean performance drop.

\setlength{\tabcolsep}{4.0pt}
    \begin{table}[!t]
        \centering
        \vspace{-3mm}
        \centering
        \resizebox{\linewidth}{!}
        {
            \begin{tabular}{ccc|cccc}
            \toprule
            \multirow{2}{4em}{Ensemble} & \multirow{2}{4em}{Weighted} & \multirow{2}{4em}{EntMin} & \multicolumn{2}{c}{Fine-grained}  & \multicolumn{2}{c}{ImageNet-X} \\
             &  &  & Acc. & Rob. & Acc. & Rob. \\
            \midrule
            \XSolidBrush & \XSolidBrush & \XSolidBrush & 53.2 & 0.4 & 44.2 & 0.3 \\
            \XSolidBrush & \XSolidBrush & \CheckmarkBold & \best{55.2} & 1.5  & \best{47.2} & 0.7 \\
            \CheckmarkBold & \XSolidBrush & \XSolidBrush & 50.6 & 40.4 & 42.7 & 29.5 \\
            \CheckmarkBold & \XSolidBrush & \CheckmarkBold & 53.3 & 44.1 & 46.7 & 34.2 \\
            \CheckmarkBold & \CheckmarkBold & \XSolidBrush & 51.6 & 43.6 & 44.8 & 33.4 \\
            \CheckmarkBold & \CheckmarkBold & \CheckmarkBold & 53.7 & \best{45.2} & 47.1 & \best{35.4} \\
            \bottomrule
            \end{tabular}
        }
    \caption{\textbf{Ablation study.} Clean and adversarial accuracies (\%) on fine-grained datasets and ImageNet dataset (CLIP-ResNet50).}
    \label{tab:ablation}
    \end{table}

\section{Conclusion}

In this paper, we for the first time explore the adversarial defense for CLIP with a test-time paradigm and propose robust test-time prompt tuning (R-TPT).
We first review the classic test time adaptation method and decompose its marginal entropy objective into a pointwise entropy term and a KL divergence term.
We find that minimizing KL divergence will introduce conflicts when meeting the adversarial test instance, thus we discard KL divergence terms and only optimize the textual prompt with the pointwise entropy.
We also introduce a reliability-based weighted ensembling strategy to utilize knowledge of the augmented views, which contain more knowledge than the risky original input under adversarial attacks.
Extensive results show that R-TPT achieves the best defense against various adversarial attacks among all baseline methods and maintains a clean adaptation performance.
We believe that our work will provide a new perspective on the defense and shed light on the safety issues of VLMs.

\section*{Acknowledgements}
This work was funded by the National Natural Science Foundation of China under Grants (62276256, U2441251) and the Young Elite Scientists Sponsorship Program by CAST (2023QNRC001).
The authors would like to thank Professor Tan Tieniu for his valuable guidance and contribution to this work.

{
    \small
    \bibliographystyle{ieeenat_fullname}
    \bibliography{main}
}

\clearpage
\maketitlesupplementary
\section{Algorithm}
Here, we provide the pseudocode algorithm of R-TPT to show the process of our proposed defense method clearly.

\begin{algorithm}[ht]
\caption{Pseudocode of R-TPT.}
\small 
\label{alg:prodding}
\begin{algorithmic}
 \STATE {\bfseries Require: Test sample $x_t$, CLIP model.}
 \STATE $\triangleright$ Augment $x_t$ via AugMix to obtain views $\{x_i\}_{i=0}^N$ and select low-entropy views $\mathcal{B}$.
 \STATE $\triangleright$ Update textual prompts via minimizing pointwise entropy of selected views $\mathcal{B}$ via Eq.4.
 \STATE $\triangleright$ Obtain the reliability $\{r_i\}_{i=0}^N$ of all views via Eq.6.
 \STATE $\triangleright$ Obtain the robust prediction by ensembling the predictions $\{p_i\}_{i=0}^N$ of all views weighted by the reliability $\{r_i\}_{i=0}^N$.
\end{algorithmic}
\end{algorithm}

\section{Datasets}
We provide the content, number of categories and number of images of all datasets involved in the experimental section in Table~\ref{tab:dataset}.

\setlength{\tabcolsep}{2.0pt}
    \begin{table}[ht]
        \centering
        \resizebox{0.48\textwidth}{!}
        {
            \begin{tabular}{llcc}
            \toprule
            Dataset & Description & \# Classes & \# Test \\
            \midrule
            Caltech101 & Object images & 100 & 2,465 \\
            Pets & Pet images & 37 & 3,669 \\
            Cars & Car images & 196 & 8,041 \\
            Flower102 & Flower images & 102 & 2,463 \\
            Aircraft & Aircraft images & 100 & 3,333 \\
            DTD & Describable textures dataset & 47 & 1,692 \\
            EuroSAT & Sentinel-2 satellite images & 10 & 8,100 \\
            UCF101 & Human action images & 101 & 3,783 \\
            \midrule
            ImageNet & Object and scene images & 1,000 & 50,000 \\
            ImageNet-A & Adversarially filtered images & 200 & 7,500 \\
            ImageNet-V2 & New test images & 1,000 & 10,000 \\
            ImageNet-R & Rendered images & 200 & 30,000 \\
            ImageNet-S & Sketch-style images & 1,000 & 50,889 \\
            \bottomrule
            \end{tabular}
        }
    \caption{Introduction of all datasets involved in experiments.}
    \label{tab:dataset}
    \vspace{-4mm}
    \end{table}

\section{Experimental Results}

\subsection{Results of Larger Backbone.}
We evaluate our method using the CLIP-ViTL/14 model \cite{radford2021learning} and present the results in Table~\ref{tab:vitl14}.
Our experiments demonstrate that R-TPT outperforms all baseline methods in terms of defense performance, highlighting its robustness even when applied to large-scale backbone architectures.
Also, we observe that, in terms of clean adaptation performance, only TPT and C-TPT exhibit positive gains, whereas the remaining methods suffer from negative transfer.

\setlength{\tabcolsep}{4.0pt}
\begin{table*}[!ht]
\vspace{-5pt}
  \centering
  \resizebox{0.98\linewidth}{!}{%
  \begin{tabular}{l|ca|ca|ca|ca|ca|ca|ca|ca|ca}
    \toprule
    \multirow{2}{*}{Method} & \multicolumn{2}{c|}{Caltech101} & \multicolumn{2}{c|}{Pets} & \multicolumn{2}{c|}{Cars} & \multicolumn{2}{c|}{Flower102} & \multicolumn{2}{c|}{Aircraft} & \multicolumn{2}{c|}{DTD} & \multicolumn{2}{c|}{EuroSAT} & \multicolumn{2}{c|}{UCF101} & \multicolumn{2}{c}{Avg.} \\
    & Acc. & \multicolumn{1}{c|}{Rob.} & Acc. & \multicolumn{1}{c|}{Rob.} & Acc. & \multicolumn{1}{c|}{Rob.} & Acc. & \multicolumn{1}{c|}{Rob.} & Acc. & \multicolumn{1}{c|}{Rob.} & Acc. & \multicolumn{1}{c|}{Rob.} & Acc. & \multicolumn{1}{c|}{Rob.} & Acc. & \multicolumn{1}{c|}{Rob.} & Acc. & \multicolumn{1}{c}{Rob.} \\
    \midrule
    CLIP \cite{radford2021learning} & 95.2 & 0.1  & 93.1 & 0.0  & 76.8 & 0.0  & 76.2 & 0.0  & 30.0 & 0.0 & 52.4 & 0.0  & \bestclean{55.1} & 0.0  & 73.7 & 0.0  & 69.1 & 0.0 \\
    Ensemble & 94.9 & 83.6 & 93.4 & 63.5 & 76.3 & 40.5 & 75.0 & 48.6 & 31.7 & 12.7 & 51.3 & 31.3 & 38.7 & 11.1 & 71.7 & 48.3 & 66.6 & 42.5 \\
    TPT \cite{shu2022test} & \bestclean{95.9} & 0.2  & 93.8 & 0.0  & 78.0 & 0.0  & \bestclean{76.9} & 0.0  & 31.6 & 0.0 & 55.1 & 0.0  & 51.8 & 0.0  & 74.7 & 0.0  & 69.7 & 0.0 \\
    C-TPT \cite{yoon2024c} & 95.6 & 0.1  & \bestclean{94.3} & 0.0  & 77.4 & 0.0  & 76.3 & 0.0  & 30.4 & 0.0 & \bestclean{55.4} & 0.0  & 54.0 & 0.0  & \bestclean{75.1} & 0.0  & \bestclean{69.8} & 0.0 \\
    MTA \cite{zanella2024test} & 95.8 & 83.1 & 93.7 & 64.9 & \bestclean{78.4} & 36.6 & 76.1 & 44.2 & \bestclean{32.7} & 8.0 & 53.4 & 27.2 & 47.8 & 7.5  & 74.7 & 47.5 & 69.1 & 39.9 \\
    R-TPT & 95.7 & \best{88.2} & 93.7 & \best{72.9} & 77.2 & \best{49.1} & 76.2 & \best{55.6} & 31.7 & \best{17.2} & 54.0 & \best{38.0} & 44.3 & \best{20.4} & 74.3 & \best{55.6} & 68.4 & \best{49.6} \\
    \bottomrule
  \end{tabular}
}
\caption{Results (\%) of various adaptation methods on \textbf{fine-grained classification datasets} with pre-trained CLIP-ViT-L/14 ($\epsilon$ = 4.0).}
\label{tab:vitl14}
\end{table*}

\subsection{Results Compared with Training-time defense Methods.}
Training-time defense methods \cite{mao2022understanding, schlarmann2024robust, li2024one} typically rely on labeled data and robust pre-trained checkpoints to achieve their performance.
To ensure a fair comparison, we have focused our main text on test-time baselines that utilize the same resources as our proposed method.
Here, we provide a comprehensive evaluation of training-time methods on fine-grained datasets in Tables~\ref{tab:training-rn50}, \ref{tab:training-vitb} to highlight the competitive performance of R-TPT, even in the absence of external data and pre-trained robust checkpoints.
It is shown that R-TPT not only remains competitive with training-time methods but also achieves significantly better performance on clean samples.
More importantly, R-TPT can further improve the robustness of training-time methods.

\setlength{\tabcolsep}{4.0pt}
\begin{table*}[!ht]
\vspace{-5pt}
  \centering
  \resizebox{0.98\linewidth}{!}{%
  \begin{tabular}{l|ca|ca|ca|ca|ca|ca|ca|ca|ca}
    \toprule
    \multirow{2}{*}{Method} & \multicolumn{2}{c|}{Caltech101} & \multicolumn{2}{c|}{Pets} & \multicolumn{2}{c|}{Cars} & \multicolumn{2}{c|}{Flower102} & \multicolumn{2}{c|}{Aircraft} & \multicolumn{2}{c|}{DTD} & \multicolumn{2}{c|}{EuroSAT} & \multicolumn{2}{c|}{UCF101} & \multicolumn{2}{c}{Avg.} \\
    & Acc. & \multicolumn{1}{c|}{Rob.} & Acc. & \multicolumn{1}{c|}{Rob.} & Acc. & \multicolumn{1}{c|}{Rob.} & Acc. & \multicolumn{1}{c|}{Rob.} & Acc. & \multicolumn{1}{c|}{Rob.} & Acc. & \multicolumn{1}{c|}{Rob.} & Acc. & \multicolumn{1}{c|}{Rob.} & Acc. & \multicolumn{1}{c|}{Rob.} & Acc. & \multicolumn{1}{c}{Rob.} \\
    \midrule
    CLIP \cite{radford2021learning} & 85.9 & 2.6  & 83.6 & 0.0  & 55.7 & 0.0  & \bestclean{61.7} & 0.0  & 15.7 & 0.0 & 40.4 & 0.8  & 23.7 & 0.0  & 59.0 & 0.0  & 53.2 & 0.4  \\
    TeCoA$^{1}$ \cite{mao2022understanding} & 78.3 & 78.3 & 76.0 & 75.8 & 22.4 & 22.3 & 33.5 & 33.4 & 5.8  & 5.8  & 26.2 & 26.0 & 16.5 & 16.6 & 38.4 & 38.1 & 37.1 & 37.0 \\
    APT$^{1}$\cite{li2024one} & 2.9  & 1.7  & 31.9 & 3.8  & 8.5  & 0.6  & 2.6  & 1.1  & 0.9  & 0.9  & 16.6 & 7.9  & 17.0 & 4.0  & 11.2 & 0.9  & 11.4 & 2.6  \\
    APT$^{1}$+TeCoA$^{1}$ \cite{li2024one} & 82.8 & \best{82.8} & 79.3 & \best{79.0} & 33.9 & 33.6 & 42.7 & 42.6 & 9.9  & 9.7  & 39.2 & \best{39.0} & \bestclean{32.9} & \best{32.9} & 51.5 & \best{51.4} & 46.5 & \best{46.4} \\
    R-TPT & \bestclean{86.7} & 79.8 & \bestclean{84.6} & 74.2 & \bestclean{58.1} & \best{42.9} & 60.6 & \best{51.9} & \bestclean{17.5} & \best{12.6} & \bestclean{41.3} & 33.5 & 21.2 & 15.9 & \bestclean{59.7} & 50.9 & \bestclean{53.7} & 45.2 \\
\bottomrule
\end{tabular}
}
\caption{Results (\%) of training-time defense methods on \textbf{fine-grained classification datasets} with pre-trained ResNet50 ($\epsilon$ = 1.0).}
\label{tab:training-rn50}
\end{table*}

\setlength{\tabcolsep}{4.0pt}
\begin{table*}[!ht]
\vspace{-5pt}
  \centering
  \resizebox{0.98\linewidth}{!}{%
  \begin{tabular}{l|ca|ca|ca|ca|ca|ca|ca|ca|ca}
    \toprule
    \multirow{2}{*}{Method} & \multicolumn{2}{c|}{Caltech101} & \multicolumn{2}{c|}{Pets} & \multicolumn{2}{c|}{Cars} & \multicolumn{2}{c|}{Flower102} & \multicolumn{2}{c|}{Aircraft} & \multicolumn{2}{c|}{DTD} & \multicolumn{2}{c|}{EuroSAT} & \multicolumn{2}{c|}{UCF101} & \multicolumn{2}{c}{Avg.} \\
    & Acc. & \multicolumn{1}{c|}{Rob.} & Acc. & \multicolumn{1}{c|}{Rob.} & Acc. & \multicolumn{1}{c|}{Rob.} & Acc. & \multicolumn{1}{c|}{Rob.} & Acc. & \multicolumn{1}{c|}{Rob.} & Acc. & \multicolumn{1}{c|}{Rob.} & Acc. & \multicolumn{1}{c|}{Rob.} & Acc. & \multicolumn{1}{c|}{Rob.} & Acc. & \multicolumn{1}{c}{Rob.} \\
    \midrule
    CLIP \cite{radford2021learning} & \bestclean{91.4} & 0.2  & \bestclean{85.1} & 0.0  & 60.1 & 0.0  & \bestclean{64.0} & 0.0  & 18.1 & 0.0 & \bestclean{43.0} & 0.0  & \bestclean{35.8} & 0.0  & 61.6 & 0.0  & \bestclean{57.4} & 0.0  \\
    TeCoA$^{4}$ \cite{mao2022understanding} & 79.3 & 78.0 & 66.9 & 63.7 & 10.2 & 9.1  & 30.8 & 28.9 & 6.6  & 5.9  & 24.5 & 24.0 & 14.5 & 14.3 & 34.6 & 33.4 & 33.4 & 32.2 \\
    FARE$^{4}$ \cite{schlarmann2024robust} & 86.3 & \best{85.4} & 76.7 & \best{73.8} & 39.2 & \best{34.4} & 37.0 & 34.0 & 9.5  & 8.5  & 28.3 & 27.3 & 16.6 & 16.3 & 44.2 & \best{41.9} & 42.2 & \best{40.2} \\
    APT$^{4}$ \cite{li2024one} & 10.7 & 0.4  & 10.0 & 0.2  & 1.5  & 0.1  & 0.9  & 0.2  & 2.6  & 0.5  & 9.0  & 0.1  & 7.8  & 6.7  & 3.7  & 0.2  & 5.8  & 1.0  \\
    APT$^{4}$+TeCoA$^{4}$ \cite{li2024one} & 81.4 & 80.2 & 66.7 & 63.9 & 20.8 & 18.9 & 42.5 & \best{40.4} & 5.2  & 5.0  & 35.2 & \best{33.7} & 29.3 & \best{29.2} & 40.2 & 39.4 & 40.2 & 38.8 \\
    R-TPT & 90.6 & 76.4 & 84.5 & 55.8 & \bestclean{63.1} & 28.4 & 62.6 & 37.6 & \bestclean{19.1} & \best{9.2} & 42.1 & 29.1 & 32.0 & 5.1  & \bestclean{62.8} & 41.0 & 57.1 & 35.3 \\
\bottomrule
\end{tabular}
}
\caption{Results (\%) of training-time defense methods on \textbf{fine-grained classification datasets} with pre-trained ViT-B/32 ($\epsilon$ = 4.0).}
\label{tab:training-vitb}
\end{table*}

\end{document}